\documentclass[times,twocolumn,final]{elsarticle}

\usepackage{medima}
\usepackage{framed,multirow}

\usepackage{amsmath,amssymb,amsfonts}
\usepackage{latexsym}
\usepackage{algorithmic}
\usepackage{url}
\usepackage{xcolor}
\usepackage{textcomp}
\usepackage{ulem}
\usepackage{multirow}
\usepackage{adjustbox}
\usepackage{subfigure}
\usepackage{soul}
\usepackage{bbm}
\usepackage{hyperref}
\definecolor{newcolor}{rgb}{.8,.349,.1}

\journal{Medical Image Analysis}

\begin{document}

\verso{Given-name Surname \textit{et~al.}}

\begin{frontmatter}

\title{Domain Generalization for Mammographic Image Analysis with Contrastive Learning}%

\author[1,2]{Zheren \snm{Li}}
\author[3]{Zhiming \snm{Cui}}
\author[2]{Lichi \snm{Zhang}}
\author[2,3]{Sheng \snm{Wang}}
\author[1]{Chenjin \snm{Lei}}
\author[1]{Xi \snm{Ouyang}}
\author[2]{Dongdong \snm{Chen}}
\author[2]{Xiangyu \snm{Zhao}}
\author[4,5]{Chunling \snm{Liu}}
\author[4,5]{Zaiyi \snm{Liu}}
\author[6]{Yajia \snm{Gu}}
\author[1,3]{Dinggang \snm{Shen}\corref{cor1}}
\cortext[cor1]{Corresponding author: 
  e-mail: Dinggang.Shen@gmail.com (D. Shen), jzcheng@ntu.edu.tw (J. Cheng)}
\author[1]{Jie-Zhi \snm{Cheng}\corref{cor1}}
  
\address[1]{Shanghai United Imaging Intelligence Co., Ltd., Shanghai 200030, China}
\address[2]{The School of Biomedical Engineering, Shanghai Jiao Tong University, Shanghai 200240, China}
\address[3]{The School of Biomedical Engineering, ShanghaiTech University, Shanghai 201210, China}
\address[4]{The Department of Radiology, Guangdong Provincial People’s Hospital, Southern Medical University, Guangzhou 510080, China}
\address[5]{Guangdong Provincial Key Laboratory of Artificial Intelligence in Medical Image Analysis and Application, Guangzhou 510080, China}
\address[6]{The Department of Radiology, Fudan University Shanghai Cancer Center, 270 Dongan Road, Shanghai 200032, China}

\received{1 May 2013}
\finalform{10 May 2013}
\accepted{13 May 2013}
\availableonline{15 May 2013}
\communicated{S. Sarkar}

\begin{abstract}
The deep learning technique has been shown to be effectively addressed several image analysis tasks in the computer-aided diagnosis scheme for mammography. The training of an efficacious deep learning model requires large data with diverse styles and qualities. The diversity of data often comes from the use of various scanners of vendors. But, in practice, it is impractical to collect a sufficient amount of diverse data for training. To this end, a novel contrastive learning is developed to equip the deep learning models with better style generalizability. Specifically, the multi-style and multi-view unsupervised self-learning scheme is carried out to seek robust feature embedding against style diversity as a pretrained model. Afterward, the pretrained network is further fine-tuned to the downstream tasks, e.g., mass detection, matching, BI-RADS rating, and breast density classification. The proposed method has been evaluated extensively and rigorously with mammograms from various vendor style domains and several public datasets. The experimental results suggest that the proposed domain generalization method can effectively improve performance of four mammographic image tasks on the data from both seen and unseen domains, and outperform many state-of-the-art (SOTA) generalization methods.
\end{abstract}


\end{frontmatter}


\section{Introduction}
\label{sec1}
Breast cancer is one of the leading causes of cancer-related deaths among women \citep{siegel2022cancer}. Mammography is the conventional imaging technique for early screening of breast cancer. It has been shown in many studies \citep{lotter2021robust,mckinney2020international,salim2020external,ouyang2021self} that advances in deep learning (DL) techniques have remarkably improved the computer-aided detection and diagnosis (CAD) schemes of mammography. The CAD schemes are composed of several components, such as lesion detection \citep{abdelrahman2021convolutional}, lesion matching \citep{yan2021towards}, malignancy classification \citep{wu2022whole}, density classification \citep{ZHAO2021103073}, and others \citep{xing2020lesion}.
However, most CAD schemes are not well tested for out-of-distribution generalization, i.e., the applicability to the unseen domains in the training process. Domain shift may result in a significant decline in performance across various vendors \citep{garrucho2022domain,liu2022medical}.

\begin{figure}[t]
\centering
\includegraphics[width=0.5\textwidth]{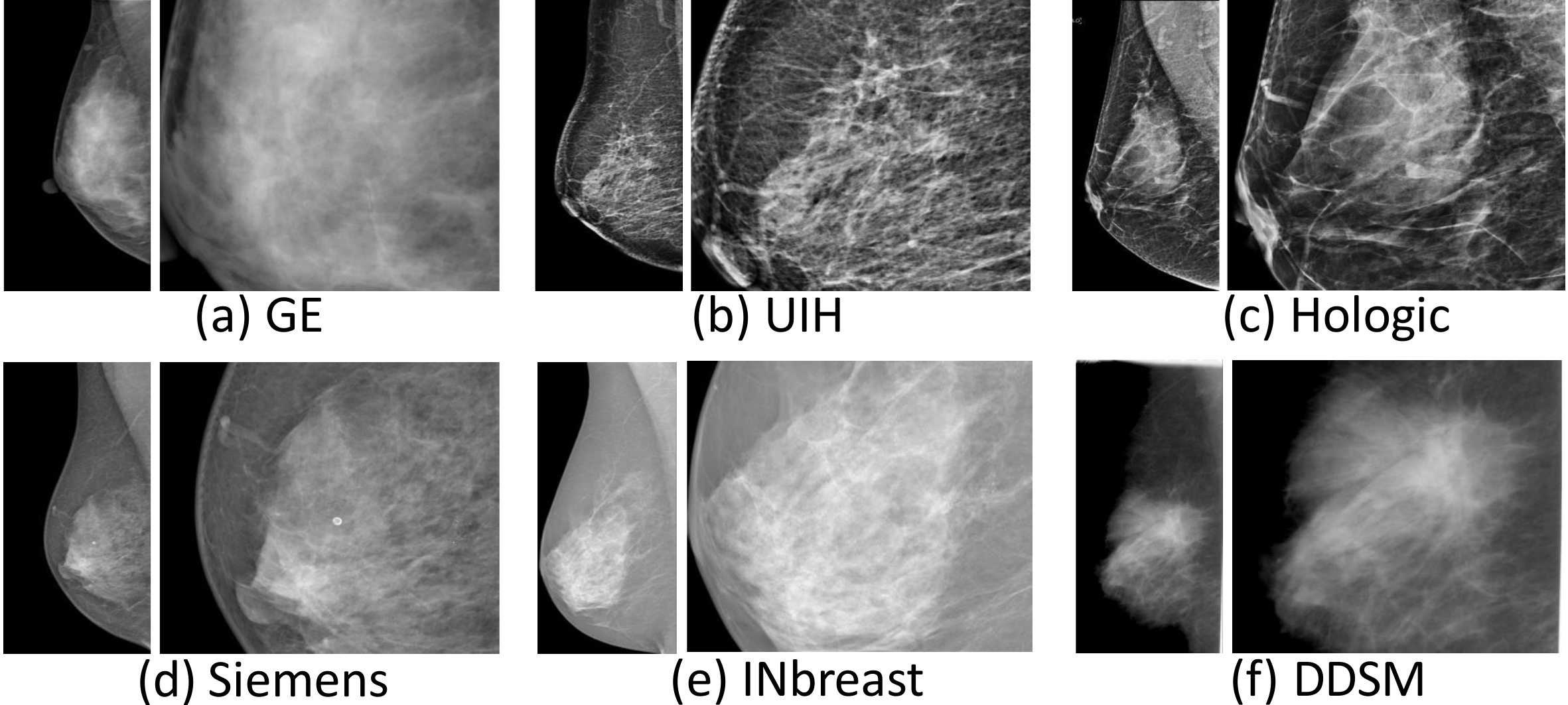}
\caption{Appearance differences across six styles/domains of mammograms: (a) GE, (b) United Imaging Healthcare (UIH), (c) Hologic, (d) Siemens, (e) INbreast, and (f) DDSM. For each style/domain, a zoomed-in of MLO view is provided to facilitate the visual comparison.}
\label{fig:different-style}
\end{figure}

As shown in Fig.~\ref{fig:different-style}, the styles of images from various vendors vary significantly. The diversity in image style among mammograms from different vendors can be attributed to various factors, such as imaging hardware and processing pipeline. With different hardware settings, the following reconstruction and post-processing algorithms may need to be specially tailored and tuned. Therefore, the style of finally generated image can appear very distinctive from vendor to vendor.

These differences in image style pose a significant challenge for CAD systems, particularly those equipped with deep learning. This is because these systems require training data that are diverse and representative of the variability in the real world. However, it is extremely expensive and impossible to gather vast amounts of diverse data from numerous vendors. Meanwhile, it is well known that domain gaps exist between datasets from different hospitals, primarily due to the differences in institutional imaging conventions and protocols. Accordingly, the generalization of DL-based CAD systems may be limited if the data of each vendor is not sufficiently included in the training stage. To overcome this challenge, a domain generalization method is needed to alleviate the burden of collecting large and diverse data from various vendors for DL-based CAD schemes.

The domain generalization for the DL technique can be classified into three categories: 1) Conventional data augmentation methods, e.g., rotation, flip, deformation, and color jittering \citep{romera2018train,volpi2018generalizing}. These augmentation methods can help the model adopt images from slightly different domains. 2) Learning-based methods with generative deep neural networks that synthesize data in the target domain~\citep{yue2019domain,kim2019diversify,zakharov2019deceptionnet}. These networks are implemented by learning an image-to-image mapping from the original to the target domain. 3) Learning-based methods for the exploration of task-specific and domain-invariant features \citep{wang2020learning,liu2020shape,dou2019domain,chen2020unsupervised}. Instead of learning an image-to-image mapping, these methods directly learn a representation-to-representation mapping.
However, all of these learning-based approaches rely on labeled data which can be expensive to acquire and is often limited in availability. It is crucial to develop a novel approach for automatically extracting domain-invariant features from large, unlabeled datasets.

To address the above-mentioned issues, here we explore the contrastive learning technique to augment the generalizability of DL-based CAD. Contrastive learning has also been shown to generate better pre-trained models for several medical image problems, e.g., diagnosis of chest radiography \citep{azizi2021big,sowrirajan2021moco} and dermatology images \citep{azizi2021big}, but is less exploited for extracting domain-invariant features. To specifically address the issue of the vendor domain gap, a multi-style and multi-view contrastive learning (MSVCL+) method is proposed to boost the CAD performance in mammography. Specifically, to attain the goal of generalization robustness to multiple vendor styles, the CycleGAN \citep{zhu2017unpaired} technique is employed to generate multiple vendor-style images from a single vendor-style image. The generated multi-style images from the same source are randomly paired as positive samples for multi-style contrastive learning (MSCL+). To further gain generalizability for the position view, the CC and MLO views of the same breast are also paired as positive samples in the scheme of multi-view contrastive learning (MVCL+). After self-supervised training, the backbone of the contrastive learning model is employed for the downstream tasks, including mass detection, multi-view mass matching, BI-RADS rating, and breast density classification. For the multi-view mass matching task, we also employ the contrastive learning approach in which the input region of interest (ROI) patches from the CC and MLO views of the same mass are treated as a positive pair. The ``+" signs of MSVCL+, MSCL+, and MVCL+ are to distinguish them from the same notations in the previous work \citep{li2021domain}.

Our main contributions are summarized in threefold:
\begin{itemize}
    \item A novel contrastive learning scheme is proposed to boost the generalizability and augment the robustness of the mammographic image analysis tasks. With style samples augmented by CycleGAN and style blending methods, the contrastive learning scheme can seek a robust feature embedding against various vendor styles. To our best knowledge, this is the first thorough study to explore the self-learning technique in addressing the mammographic domain gap problem w.r.t. vendor styles and views.
    \item The proposed method has been shown to be helpful in boosting the domain generalizability for the four mammographic tasks, namely mass detection, matching, BI-RADS rating, and breast density classification. The performances of the four tasks have been extensively and thoroughly evaluated on both seen and unseen domains.
    \item A relatively large dataset, 27,000 unlabeled and 2,700 labeled images, was involved in the development and testing of the contrastive learning and the deep learning schemes for the mammographic tasks. The proposed method is trained on three seen vendor domains and evaluated on data from six vendor domains, including seen and unseen domains. In particular, two public datasets of unseen domains were collected from populations with ethics distinctive from the in-house data. Meanwhile, a rigorous data-hungry experiment is conducted to illustrate that the proposed domain generalization method may promise better task performance than other compared SOTA methods. 
\end{itemize}

This study is an extension work of \citep{li2021domain} with significant improvements. These improvements can be summarized in threefold.
(1) Previous work \citep{li2021domain} employed CycleGAN to synthesize discrete vendor styles. Therefore, the scope of vendor styles may be finite for contrastive learning. To further enrich the diversity of the vendor styles, a simple blending method is adopted to augment the variation of synthesized styles. Higher variation of styles may offer more search space for contrastive learning to seek better feature embedding.
(2) In this study, the bundling strategy for the positive and negative pairs in contrastive learning is revised based on the knowledge of mammography. The strategy may better help the contrastive learning to find a more robust feature embedding.
(3) The size of labeled data is expanded by 75\%, encompassing a larger training and validation dataset, along with the addition of a new unseen testing dataset, while the proposed method is also subjected to more thorough and rigorous evaluation across three additional downstream tasks.

The rest of the paper is organized as follows: Section II gives a brief review of related works. Section III demonstrates the details of the proposed domain generalization method. Section IV describes experimental results and visualizations. Finally, Section V concludes the paper.

\section{Related Works}
\subsection{Domain Generalization}
Domain generalization is an important research area to overcome performance degradation caused by cross-domain variations in medical image analysis. Many methods \citep{zhang2020generalizing,liu2020shape,dou2019domain,mahajan2021domain,li2020domain} have been proposed to improve generalizability through data augmentation or exploiting the general learning strategy. However, learning-based methods are more desirable and related to our work, we focus on this category in the following.

With the recent developments in machine learning techniques, plenty of works utilize representation learning to address the domain generalization problem. They can be mainly divided into four categories: kernel-based methods \citep{blanchard2021domain,hu2020domain}; domain adversarial learning \citep{gong2019dlow,zhao2020domain}; explicit feature alignment \citep{motiian2017unified,pan2018two,fan2021adversarially}; and invariant risk minimization \citep{ahuja2021invariance,krueger2021out}. Explicit feature alignment aims to align the features from source domains to learn domain-invariant representations. For example, Motiian et al. \citep{motiian2017unified} develop a cross-domain contrastive loss for representation learning in which mapped domains are semantically matched, while remaining maximally separated. Pan et al. \citep{pan2018two} implement instance normalization layers to CNNs to improve model generalization. Recently, Fan et al. \citep{fan2021adversarially} present that adaptively learning the normalization with the combination of multiple normalization strategies can improve domain generalization capabilities.

Following the success of self-supervised learning \citep{jing2020self}, several studies\citep{carlucci2019domain,kim2021selfreg,jeon2021feature} build self-supervised tasks from large-scale unlabeled data to learn generalized representations. For instance, Carlucci et al. \citep{carlucci2019domain} present a self-supervision task of solving jigsaw puzzles to learn the concepts of spatial correlation. Daehee et al. \citep{kim2021selfreg} propose a self-supervised contrastive regularization approach that utilizes only positive data pairs. Seogkyu et al. \citep{jeon2021feature} exploit stylized features for regularization via consistency loss and
domain-aware supervised contrastive loss. However, these methods were not specifically designed for medical images. Therefore, the efficacy of these methods \citep{carlucci2019domain,kim2021selfreg,jeon2021feature} for medical image analysis is unknown.

\subsection{Contrastive Learning}
Recently, contrastive learning has been proven to be notably effective in self-supervised learning. It draws samples from the same class (a positive pair) close together while drives different samples (or negative pairs) apart in the latent embedding space through contrastive loss. 
Chen et al. \citep{chen2020simple} propose SimCLR, a contrastive learning-based system for learning effective presentations by maximizing the similarity between two different augmented views from the original image. He et al. \citep{he2020momentum} utilize momentum contrast (MoCo) which divides each image into a query, and then generates a key by performing two distinct augmentations. MoCo v2 \citep{chen2020improved} incorporates enhancements such as the integration of a two-layer MLP head with ReLU during the unsupervised training stage, and the implementation of a data augmentation technique involving blurring.
SimCLR and MoCo provide promising results that are much closer to supervised-learning compared to other self-supervised learning methods. These frameworks are widely used in the pre-train stage and showed constant improvement for various downstream tasks. Later, Grill et al. \citep{grill2020bootstrap} introduce a technique called BYOL for learning feature representations without the massive number of negative pairs. Basically, BYOL added another MLP on the SimCLR to create an asymmetrical architecture.
In this paper, we adopt the SimCLR as the pre-training method, because of its advantage of easy implementation. Meanwhile, we further replace the SimCLR with both MoCo v2 and BYOL as the pretraining methods. The experimental results suggest there is no significant difference among all pre-training methods.

\begin{figure}[!t]
\centering
\includegraphics[width=0.48\textwidth]{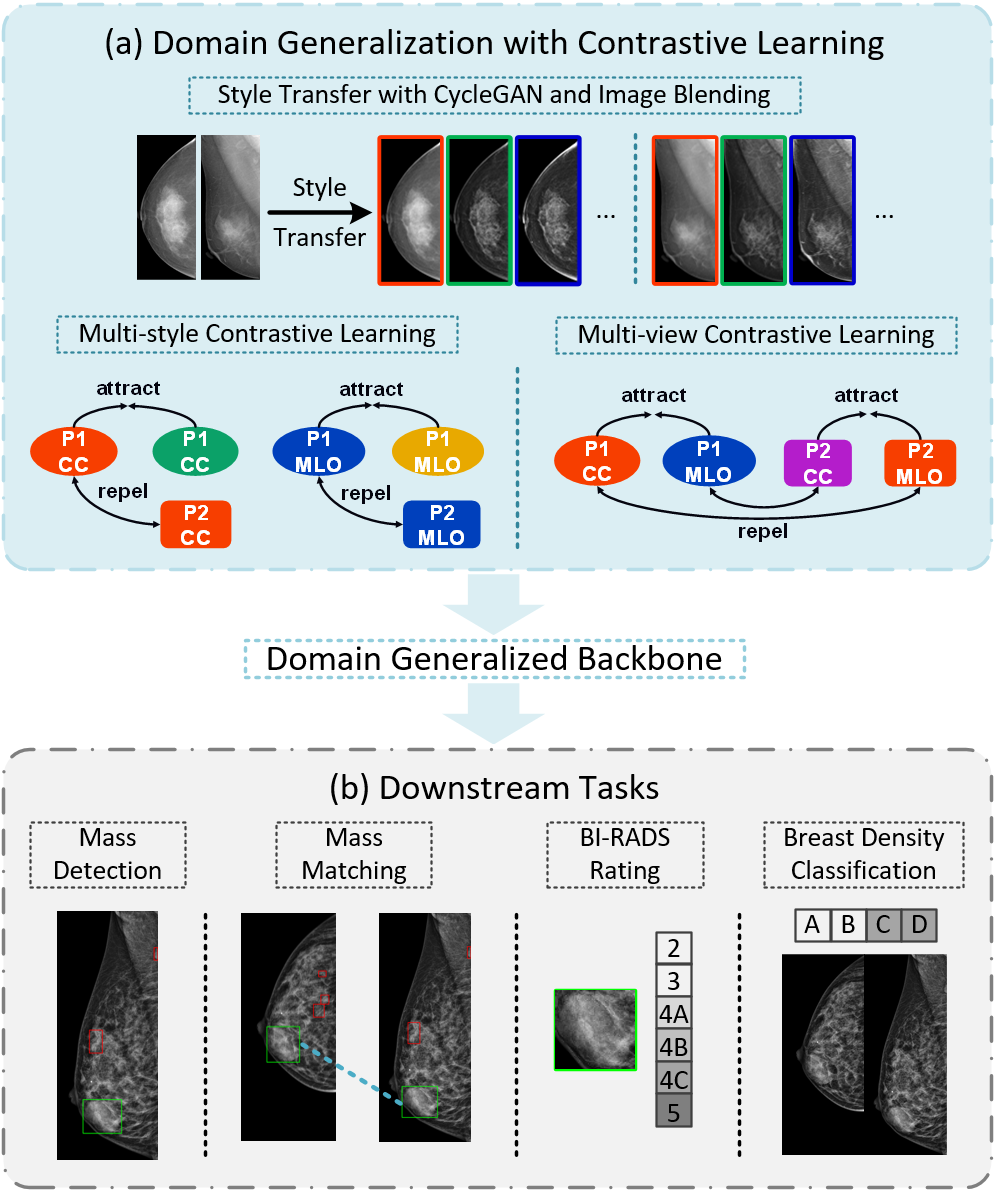}
\caption{Our domain generalization method for mammograms involves two main steps. In stage (a), the CycleGAN augmented with an image blending technique is adopted to generate diverse vendor styles. The contrastive learning scheme is further applied to learn better feature embedding against various domains, from the generated styles. In stage (b), the feature embedding encoded in the backbone of the contrastive learning model serves as the pre-trained model for downstream tasks.
}
\label{fig:overview} 
\end{figure}

\begin{figure*}[!t]
  \centering
  \includegraphics[width=0.95\textwidth]{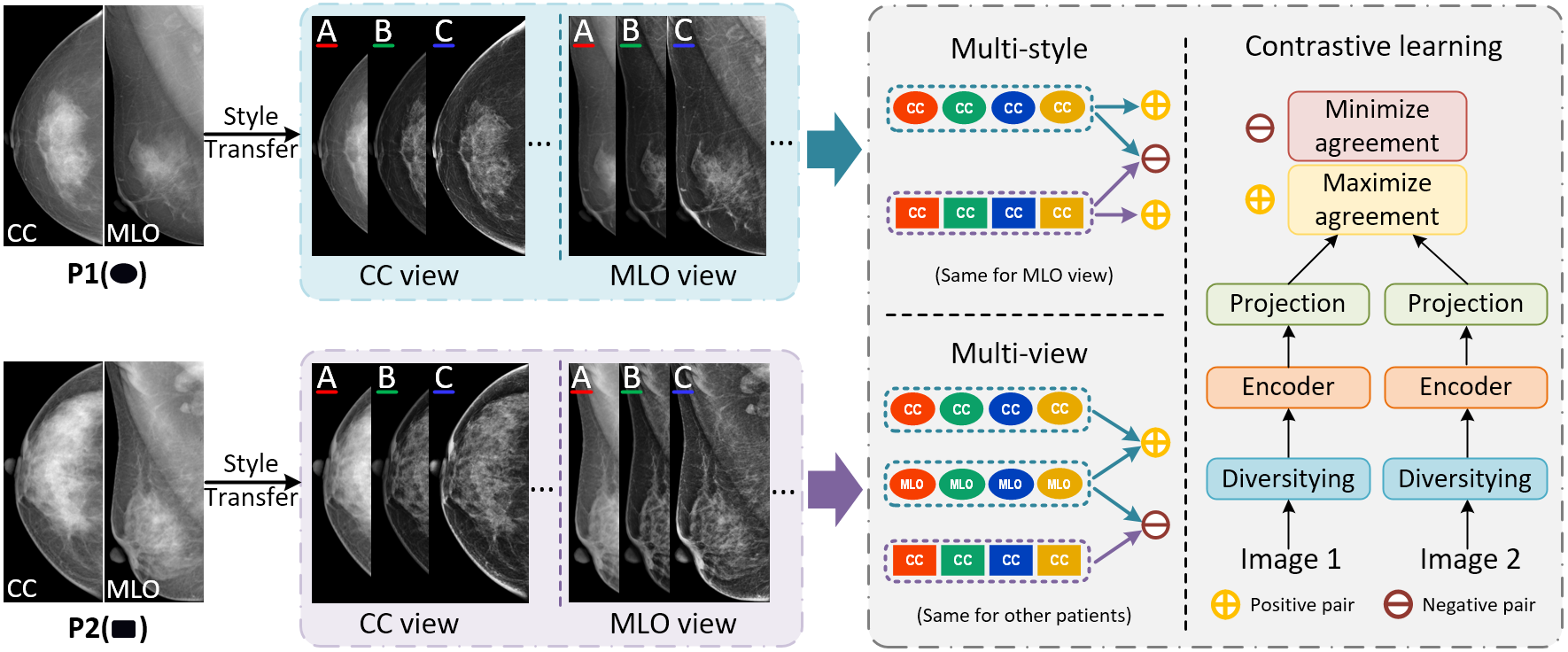}
  \caption{Illustration of the multi-style and multi-view contrastive learning scheme. P1 and P2 indicate different patients. The letters A, B, and C represent different vendor styles generated from one source domain. The right box shows the concept of bundling positive and negative pairs for the contrastive learning schemes. Specifically, the red circle with a minus sign inside stands for the negative pair, while the yellow circle with a plus sign inside indicates the positive pair.}
  \label{fig:msvcl}
\end{figure*}

\subsection{Mammographic Image Analysis}
Mass detection is one of the most fundamental problems in mammographic image analysis\citep{abdelrahman2021convolutional}. Jung et al. \citep{jung2018detection} use the RetinaNet \citep{lin2017focal} model as a one-stage mass detector in mammograms. Shen et al. \citep{shen2020unsupervised} propose a framework that depends on adversarial learning to detect the mass in mammograms. The adversarial learning helps align the latent target features from unlabeled datasets with labeled source domain latent features. 
Moreover, CAD schemes also require supplementary mammographic image analysis functions such as mass matching, malignancy classification, and breast density classification. Yan et al. \citep{yan2021towards} exploit multi-tasking properties of deep networks to jointly learn mass matching and classification. Yang et al. \citep{yang2021momminet} present MommiNet-v2 to incorporate the malignant information from both biopsies and BI-RADS categories. Zhao et al. \citep{ZHAO2021103073} present an innovative bilateral-view adaptive spatial and channel attention network (BASCNet) for fully automated breast density classification. 
Despite the many existing methods for mammographic image analysis, few of them have been evaluated on unseen domains, which means that their out-of-distribution generalization has not been thoroughly examined.

\section{Method}
Fig.~\ref{fig:overview} illustrates the proposed domain generalization framework for mammography analysis. Our approach consists of two main stages. In stage (a), a style-robust backbone is trained with the contrastive learning technique as the pre-trained model for the downstream tasks. To facilitate contrastive learning, the CycleGAN technique is first employed to diversify the vendor styles. In the following, a multi-style and multi-view contrastive learning scheme is carried out to embed a general feature space, which is more robust to both the vendor-style and view domains, in the pre-trained backbone. The common downstream tasks of mammography analysis, including mass detection, matching, BI-RADS rating, and breast density classification, are further fine-tuned with the pre-trained backbone for better generalizability.

\subsection{Contrastive Learning Scheme}
Contrastive learning is a self-supervised learning method that trains a network to encode the image representation into a proper vector space without the requirement of explicit annotations. The derived model from contrastive learning commonly serves as a pre-trained model for the training of various downstream tasks.

The basic idea of contrastive learning is to pack the diversified images of the same class/object/subject as positive pairs for the exploration of proper feature embedding. Specifically, given a mini-batch of $N$ images, each example is randomly augmented twice with diversifying operations, e.g., cropping and rotation, to generate an augmented mini-batch with $2N$ samples. In the augmented mini-batch, two samples from the same image source are treated as a positive pair $(i,j)$, whereas the other $2(N-1)$ samples within the mini-batch are regarded as negative pairs. With the positive and negative pairs, contrastive learning is driven by the contrastive loss to maximize the agreement for the positive pairs. The contrastive loss is defined as:

\begin{equation}
\ell_{i,j} = -\log \frac{{\rm exp}({\rm sim}(z_i,z_j)/\tau)}{\sum_{k=1}^{2N} \mathbbm{1}_{[k\neq i]}{\rm exp}({\rm sim}(z_i,z_k)/\tau)},
\label{eq:contrastive_loss}
\end{equation}
where ${\rm sim}(\cdot)$ is the dot product and $z$ refers to the extracted features. $\mathbbm{1}_{[k\neq i]} \in \{0,1\}$ is an indicator function equaling 1 when $k\neq i$. $\tau$ is a temperature parameter. 

By maximizing the agreement for the positive pairs, the learned features of corresponding images are supposed to “attract” each other, while the learned features of non-corresponding images “repel” each other by minimizing the agreement for the negative pairs. We further leverage the concept of contrastive learning to explore the generalization of various vendor styles and the domains of CC and MLO views. The details of multi-style and multi-view contrastive learning are elaborated in the following section.

\begin{figure}[!t]
  \centering
  \includegraphics[width=0.49\textwidth]{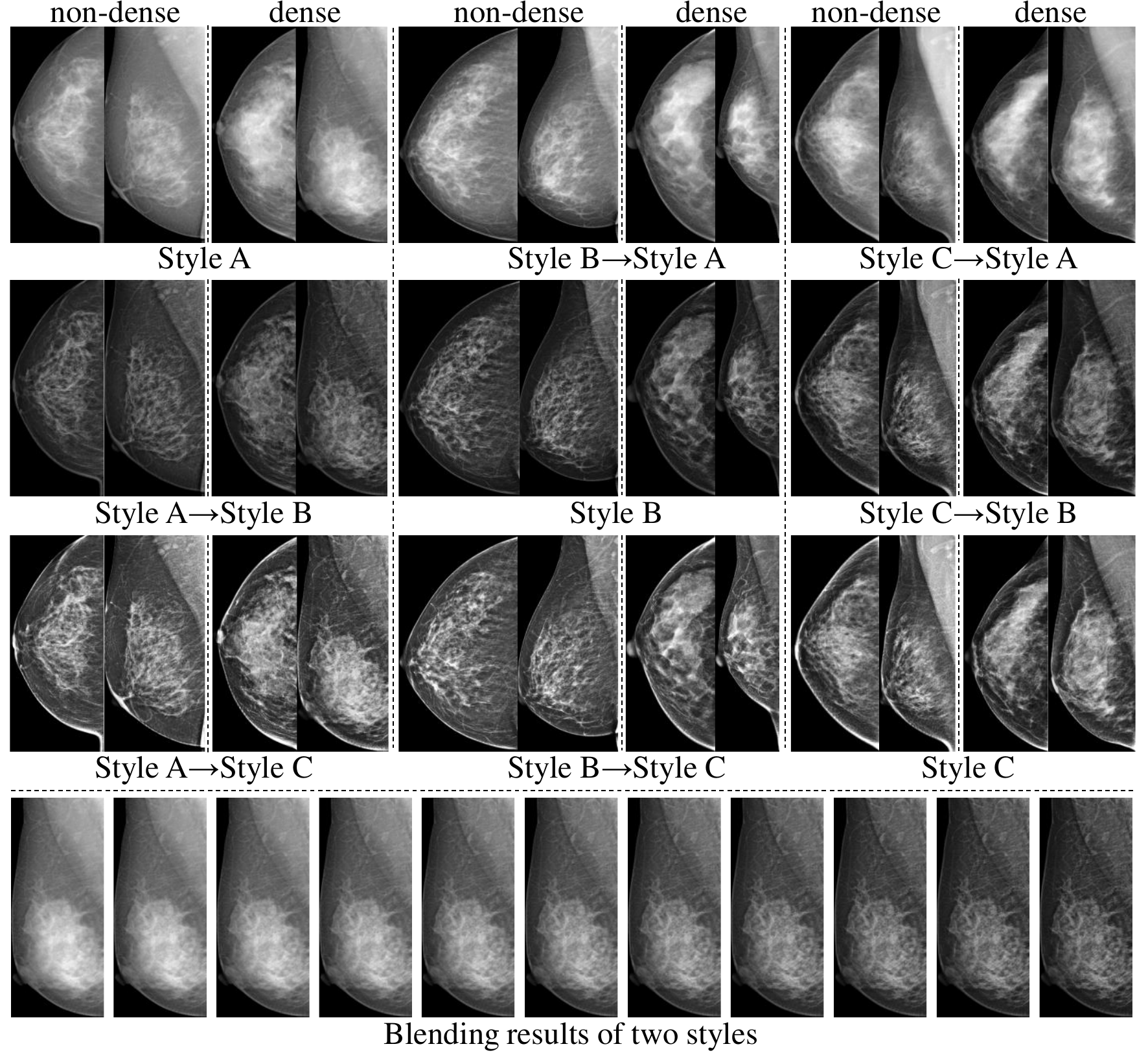}
  \caption{Visual comparison of style transfer results. The first three rows display the results of style transfer using original images from style A, B, and C. The left column displays generated outcomes of A$\rightarrow$B and A$\rightarrow$C, whereas the center column represents generated outcomes of B$\rightarrow$A and B$\rightarrow$C. The right column presents the generated results of C$\rightarrow$A and C$\rightarrow$B. Each style contains both non-dense and dense breast cases. The last row demonstrates the series of synthesized styles from styles A and B with blending method shown in Eq.~\ref{eq:img_blending}.
  }
  \label{fig:style_transfer} 
\end{figure} 

\subsection{Multi-style and Multi-view Contrastive Learning}  
To seek feature embedding with better generalizability for the vendor-style and view domains, a multi-style and multi-view contrastive learning approach is devised. The synergy of the two learning schemes is illustrated in Fig. \ref{fig:msvcl}. Specifically, CycleGAN is employed to diversify the vendor styles of the original data. The style diversification is further augmented with an image blending operator. Afterward, the positive and negative pairs are packaged for contrastive learning in the multi-style and multi-view schemes, respectively. The synergized contrastive learning scheme is driven by the optimization of both multi-style and multi-view contrastive losses.

\subsubsection{Multi-style Contrastive Learning} 
The image styles vary for different vendors, see Fig. \ref{fig:different-style}. Therefore, the various vendor styles are regarded as distinctive domains. To endow the backbones of downstream tasks with better generalizability, we exploit the CycleGAN \citep{zhu2017unpaired} augmented with a blending technique as the diversifying operation for contrastive learning. Specifically, given $M$ seen vendor-style domains, we train $\tbinom{M}{2}$ generators, which map the data distribution of source domain $\Omega_i$ to target $\Omega_j$ domain, $\forall i, j \in M$.
With the $\tbinom{M}{2}$ generators, each of the original $N$ images in the training set can be diversified into $M-1$ transferred images, which is illustrated in the first three rows of Fig.~\ref{fig:style_transfer}. Moreover, we further use the image blending technique to achieve a smooth transition between two styles:
\begin{equation}
style_{blend} = style_{A} \times (1.0-alpha) + style_{B} \times alpha,
\label{eq:img_blending}
\end{equation}
where $alpha$ is the interpolation factor randomly selected from a set of values ranging from 0 to 1 with an interval of 0.1. The blending results are illustrated in the last row of Fig.~\ref{fig:style_transfer}. In practice, for each original image, the samples for the bundling of positive and negative pairs are randomly picked from the pool of $L = \tbinom{M}{2} \times 9 + M$ blended style. With the aid of this blending technique, the diversity of the style samples for the MSCL+ will be significantly enhanced. Compared to the previous MSCL, the number of style samples has increased by 10 times.

In MSCL+, the two styles of the same source image (e.g., style A and style B from the CC view of a patient) are attracted together, see stage (a) in Fig.~\ref{fig:overview}, while the same view of different patients (e.g., CCs of two patients, regardless of styles) are repelled from each other. The positive pairs for contrastive learning are constituted with any two images diversified from the same source image in the original $N$ image set. Therefore, there are possible $N \times \tbinom{L}{2}$ positive pairs available for random selection in contrastive learning. Referring to the right box in Fig.~\ref{fig:msvcl}, the bundling strategy for negative pairs is further refined by only considering the same view positions from different patients, e.g., CC views from patient 1 and patient 2, to exclude the factor of domain gap between distinct views. The combination of CC and MLO views from different patients for a negative pair, which were involved in previous MSCL, may confuse contrastive learning and not be informative samples for effective training. The contrastive learning is then carried out by minimizing the Eq.~\ref{eq:contrastive_loss} to seek feature embedding space with better generalizability to various vendor domains.


\subsubsection{Multi-view Contrastive Learning}
A standard examination of mammography consists of CC and MLO views for each breast. Because the two standard views are taken from different angles of the same breast, they are mutually complementary for diagnosis. To seek domain-invariant feature embedding against different view domains, we explore the contrastive learning scheme to consider the distinctive view domains. Specifically, the CC and MLO views of the same breast from the same patient (e.g., LCC and LMLO of a patient) are treated as a positive pair, whereas the other combination of the CC and MLO views from different patients (e.g., LCC of patient 1 and LMLO of patient 2) is a negative pair. 
Similarly to MSCL+, the bundling strategy for negative pairs is further restricted to exclude the pairing of the same view positions from different patients, e.g., LCC views of patients 1 and 2, in MVCL+. This type of pairing was adopted in previous MVCL and may be redundant for training. To further enrich the sample diversity, we implement style-diversifying operations for the CC and MLO in each positive or negative pair. With the prepared sample pairs, MVCL+ can be carried out for the embedding of view-invariant features. The outperformance of MSCL+ and MVCL+ to the previous versions will be shown in the ablation experiments.

\begin{table*}[!t]
\caption{Distributions of data usage for different stages. 
}\label{tab:dataset}
\centering
{\fontsize{9}{11}\selectfont 
\begin{tabular}{c|c|c|c|c|c}
\hline
Domain & Dataset & Vendor & Style Transfer & Self-Supervision & Downstream Tasks (train/val/test) \\ \hline
& Style A & GE & 1000 & 8000 & 600/100/100 \\
Seen & Style B & UIH & 1000 & 8000 & 600/100/100 \\
& Style C & HOLOGIC & 1000 & 8000 & 600/100/100 \\
\hline
& Style D & SIEMENS & 0 & 0 & 0/0/100 \\
Unseen & Style E & INbreast & 0 & 0 & 0/0/100 \\
& Style F & DDSM & 0 & 0 & 0/0/100 \\
\hline
\end{tabular}
}
\end{table*}

\subsection{Downstream tasks}
\subsubsection{Mass Detection}
In this study, the classic single-view detection network of FCOS \citep{tian2020fcos} is employed to identify mass in mammography. The derived pre-trained model from the self-supervised learning stage is adopted as the backbone of the FCOS architecture for the mass detection tasks. The detected results from CC and MLO paired images are two sets of candidate bounding boxes: $B_{cc} = \{b_{cc}^1,...,b_{cc}^i,...,b_{cc}^N\}$ and $B_{mlo} = \{b_{mlo}^1,...,b_{mlo}^j,...,b_{mlo}^M\}$. 

\begin{figure}[!t]
  \centering
  \includegraphics[width=0.45\textwidth]{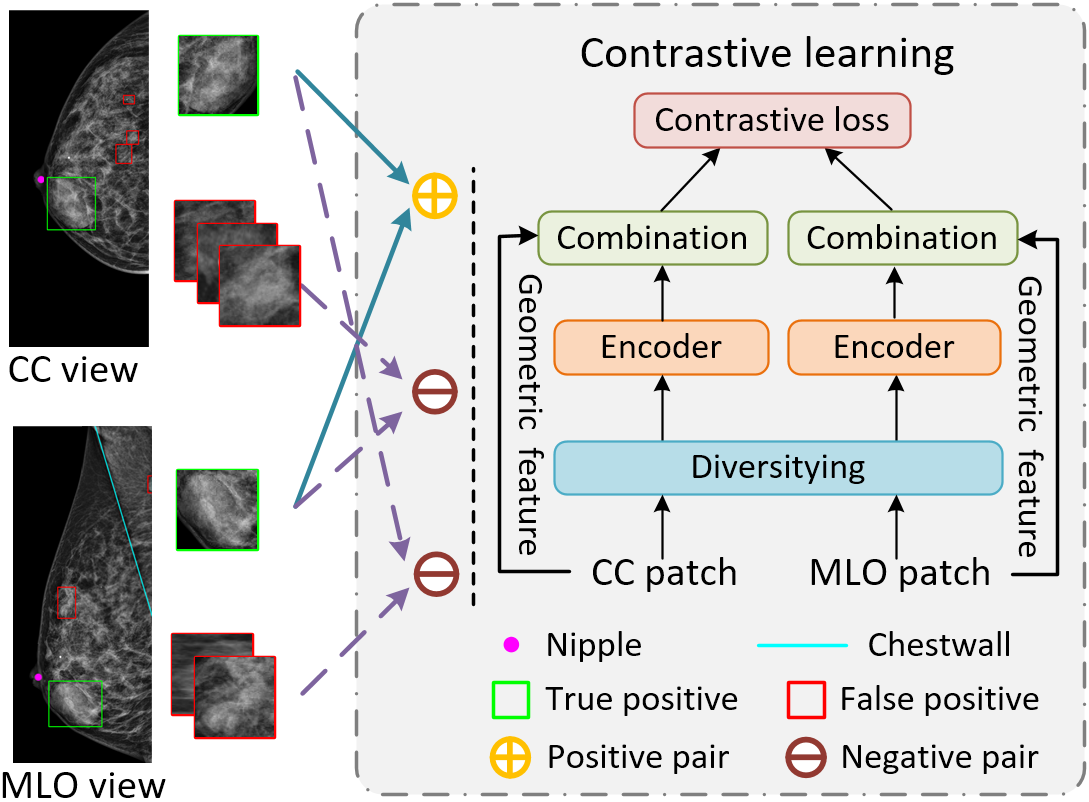}
  \caption{The multi-view mass matching scheme with contrastive learning. The green boxes indicate regions of true positive mass, while the red boxes refer to false positives. In addition, the purple dots represent the location of the nipple, and the blue line suggests the chestwall.}
  \label{fig:matching} 
\end{figure}

\subsubsection{Mass Matching}
In the clinical reading, two found mass instances in the CC and MLO views are regarded as the same mass if several anatomical metrics, e.g., distance to nipple, appearance, and shape, are satisfied. If a mass can be found in both views, the diagnostic confidence is higher. On the other hand, the contrastive learning technique naturally takes two inputs for pair matching and can potentially be applied to realize the mass matching process. Inspired by this, we address mass matching with the incorporation of anatomical cues in contrastive learning. Specifically, the two sets of true-positive (TP) and false-positive (FP) bounding boxes from the FCOS are defined based on the IoU metrics with the ground truth, and these boxes can be further named as $TP_{cc}$, $FP_{cc}$, $TP_{mlo}$ and $FP_{mlo}$ w.r.t. MLO and CC views. The pair of $TP_{cc}$ and $TP_{mlo}$ that satisfies anatomical metrics is treated as a positive pair with the matching label Y of 1. On the other hand, the combination of boxes with any one from either $FP_{cc}$ or $FP_{mlo}$, is regarded as a negative pair with label $Y$ of 0.

As shown in Fig. \ref{fig:matching}, the CC/MLO patch matching is based on the metrics of appearance similarity, distances to the nipple and chestwall, and the box size. In the training process, the max margin contrastive loss is adopted as the matching error:
\begin{equation}
L_{mat}=\frac{1}{2K}\sum_{k=1}^{K}YD^{2}+(1-Y)max(m-D,0)^{2},
\label{eq:contrastive_loss_match} 
\end{equation}
where $D = \left \| f_{1} - f_{2} \right \|_{2} $ refers to the L2 distance between the embedding features of paired samples. The label $Y $ is 1 if the two samples are matched, otherwise 0. The margin $m$ is a hyper-parameter, which suggests the lower bound distance between patches that are not matched. 

\subsubsection{BI-RADS Rating and Breast Density Classification}
In addition to mass detection and matching, we also illustrate the domain generalization performance on the two major downstream classification tasks of mammography analysis, i.e., BI-RADS rating and breast density classification. For BI-RADS rating, we adopt the pre-trained model for the classification model with five BI-RADS scores of 2$\sim$3, 4A, 4B, 4C, and 5. In the context of breast density classification, the pre-trained model is also adopted for the classification model with the input of a whole mammogram, either CC or MLO view. The breast density classes are non-dense and dense, i.e., the density categories of A$\sim$B and C$\sim$D.

\begin{table*}[!t]
\caption{Ablation analysis w.r.t. various pre-training settings for mass detection tasks. The assessment metric is mAP[\%].}\label{tab:ablation_detection}
\centering
{\fontsize{9}{11}\selectfont 
\begin{adjustbox}{center}
\begin{tabular}{c|c|c|c|c|c|c|c|c|c}
\hline
\multirow{2}{*}{Pre-training Method} & \multirow{2}{*}{Pre-training Dataset} &\multicolumn{4}{c|}{Seen domain}&\multicolumn{4}{c}{Unseen domain} \\
\cline{3-10} & & Style A & Style B & Style C & Avg. & Style D & Style E & Style F & Avg. \\ \hline
Random &  {None}                            & 59.8          & 70.5 & 66.5 & 65.6$\pm$5.4            & 51.3 & 79.9 & 65.8 & 65.7$\pm$14.3 \\
Supervised &  {ImageNet}                    & 84.0          & 85.7 & 79.1 & 82.9$\pm$3.4            & 82.4 & 76.3 & 81.0 & 79.9$\pm$3.2 \\
SimCLR & {Mammo}                            & 84.4          & 91.6 & 84.1 & 86.7$\pm$4.2            & 82.2 & 77.3 & 75.0 & 78.2$\pm$3.7\\
SimCLR & {$\rm ImageNet\rightarrow Mammo$}  & 86.9          & 92.0 & 86.0 & 88.3$\pm$3.2    & 86.2 & 80.5 & 84.2 & 83.6$\pm$2.9 \\
\hline
MSCL & {$\rm ImageNet\rightarrow Mammo$}    & 87.6          & 92.3 & 85.8 & 88.6$\pm$3.4            & 88.5 & 85.2 & 86.0 & 86.6$\pm$1.7\\
MVCL & {$\rm ImageNet\rightarrow Mammo$}    & 88.9          & 92.4 & 84.3 & 88.5$\pm$4.1            & 86.7 & 85.3 & 87.3 & 86.4$\pm$1.0\\
MSVCL & {$\rm ImageNet\rightarrow Mammo$}   & 91.8          & 94.0 & \textbf{89.1} & \textbf{91.6$\pm$2.5} & 87.3 & \textbf{89.7} & 88.4 & 88.5$\pm$1.2\\
\hline
MSCL+ & {$\rm ImageNet\rightarrow Mammo$}   & 92.2          & \textbf{94.3} & 86.7 & 91.1$\pm$3.9   & 89.5 & 86.2 & 89.0 & 88.2$\pm$1.8\\
MVCL+ & {$\rm ImageNet\rightarrow Mammo$}   & 90.8          & 94.1 & 86.7 & 90.5$\pm$3.7            & 89.7 & 86.7 & 90.4 & 88.9$\pm$2.0\\
MSVCL+ & {$\rm ImageNet\rightarrow Mammo$}  & \textbf{92.9} & 93.8 & 87.0 & 91.2$\pm$3.7    & \textbf{90.4} & 89.4 & \textbf{94.1} & \textbf{91.3$\pm$2.5}\\
\hline
\end{tabular}
\end{adjustbox}
}
\end{table*}

\begin{table*}[!t]
\caption{Ablation analysis w.r.t. various pre-training settings for the four tasks. The assessment metrics are mAP[\%] for mass detection, and acc[\%] for the tasks of mass matching, BI-RADS rating, and breast density classification.}\label{tab:ablation}
\centering
{\fontsize{9}{11}\selectfont 
\begin{adjustbox}{center}
\begin{tabular}{c|c|c|c|c|c|c|c|c|c}
\hline
\multirow{2}{*}{Pre-training Method} & \multirow{2}{*}{Pre-training Dataset} &\multicolumn{4}{c|}{Seen domain}&\multicolumn{4}{c}{Unseen domain} \\
\cline{3-10} &                              & Detection & Match & BI-RADS & Density & Detection & Match & BI-RADS & Density \\ 
\hline
Random & {None}                             & 65.6$\pm$5.4      & 75.0$\pm$1.4    & 55.1$\pm$3.9   & 65.3$\pm$1.2 & 65.7$\pm$14.3      & 60.3$\pm$4.7      & 66.1$\pm$34.7 & 40.3$\pm$12.6 \\
Supervised & {ImageNet}                     & 82.9$\pm$3.4      & 86.3$\pm$5.8    & 85.4$\pm$3.5   & 90.3$\pm$2.1 & 79.9$\pm$3.2      & 87.4$\pm$3.4      & 81.4$\pm$17.1 & 77.0$\pm$6.6 \\
SimCLR & {Mammo}                            & 86.7$\pm$4.2      & 79.7$\pm$3.7    & 79.4$\pm$4.7   & 92.3$\pm$1.2 & 78.2$\pm$3.7      & 81.5$\pm$7.6      & 79.5$\pm$17.2 & 73.0$\pm$19.2 \\
SimCLR & {$\rm ImageNet\rightarrow Mammo$}  & 88.3$\pm$3.2      & 80.0$\pm$6.7    & 87.4$\pm$6.7   & 91.7$\pm$2.5 & 83.6$\pm$2.9      & 85.0$\pm$7.4      & 83.3$\pm$18.4 & 76.3$\pm$16.3 \\
\hline
MSCL & {$\rm ImageNet\rightarrow Mammo$}    & 88.6$\pm$3.4      & 85.7$\pm$4.1    & 86.4$\pm$3.1   & 91.7$\pm$1.5 & 86.6$\pm$1.7      & 87.1$\pm$6.6      & 85.8$\pm$13.5 & 80.3$\pm$6.2 \\
MVCL & {$\rm ImageNet\rightarrow Mammo$}    & 88.5$\pm$4.1      & 89.3$\pm$2.9    & 89.2$\pm$4.8   & 90.7$\pm$0.6 & 86.4$\pm$1.0      & 84.6$\pm$7.1      & 84.2$\pm$18.3 & 80.3$\pm$6.6 \\
MSVCL & {$\rm ImageNet\rightarrow Mammo$}   & \textbf{91.6$\pm$2.5} & 89.6$\pm$4.3 & 89.3$\pm$4.9  & 91.7$\pm$3.2 & 88.5$\pm$1.2      & 91.5$\pm$2.9      & 85.9$\pm$9.4 & 81.0$\pm$4.4 \\
\hline
MSCL+ & {$\rm ImageNet\rightarrow Mammo$}  & 91.1$\pm$3.9  & \textbf{92.2$\pm$1.9} & \textbf{90.0$\pm$4.5} & \textbf{93.0$\pm$1.0} & 88.2$\pm$1.8 & 89.7$\pm$4.2 & 86.8$\pm$9.4 & 82.3$\pm$5.8 \\
MVCL+ & {$\rm ImageNet\rightarrow Mammo$}  & 90.5$\pm$3.7      & 89.0$\pm$7.0    & 87.6$\pm$3.5   & 92.3$\pm$1.5   & 88.9$\pm$2.0     & 86.9$\pm$5.9     & 86.7$\pm$14.8 & 83.3$\pm$5.6 \\
MSVCL+ & {$\rm ImageNet\rightarrow Mammo$} & 91.2$\pm$3.7      & 90.6$\pm$3.1    & 89.0$\pm$5.5   & 92.0$\pm$2.6   & \textbf{91.3$\pm$2.5} & \textbf{92.9$\pm$2.5} & \textbf{87.8$\pm$9.3} & \textbf{84.3$\pm$8.3}\\
\hline
\end{tabular}
\end{adjustbox}
}
\end{table*}

\section{Experiments and Results}
\label{sec:experiments}
\subsection{Datasets}
Table~\ref{tab:dataset} shows the details of data usage in this paper. One in-house and two public datasets are involved for model training and performance evaluation. The in-house dataset was collected from machines of four vendors: GE, United Imaging Healthcare (UIH), Hologic, and Siemens, denoted as A, B, C, and D, respectively, for short. All image data from four vendors were acquired from Asian women. The data from vendors A, B, and C are set as seen domains, whereas vendor D is treated as an unseen domain. The annotation for the in-house dataset was first carried out by two radiologists with 3-5 years of experience and then reviewed by a senior radiologist with more than 10 years of experience. To further evaluate the generalizability of the proposed method, the two public datasets, i.e., INbreast \citep{moreira2012inbreast} and DDSM \citep{DDSM2000}, denoted as E and F, respectively, are treated as the unseen datasets. In total, 29,700 mammograms (14,850 CC/MLO pairs) are involved in this study. 27,000 unannotated images are used for the training of stage (a) with style transfer and contrastive learning schemes. Specifically, all of the unannotated images are from the seen domain, including styles A, B, and C. The number of images used in each style is 1000 for style transfer and 8000 for self-supervision, respectively. The remaining 2,700 annotated images are employed for the training, validation, and testing of stage (b), i.e., the downstream tasks. For each style in the seen domain, there are 600, 100, and 100 images for training, validation, and testing, respectively. For each style in the unseen domain, there are 100 images for testing. Datasets A, B, C, D, and E are tested on four downstream tasks, while dataset F is tested on three downstream tasks without BI-RADS rating, because the BI-RADS scores are not available in the DDSM.

\subsection{Implementation Details}
In stage (a), the generator of the CycleGAN is a ResNet with 9 blocks of 20 convolution layers, whereas the discriminator is PatchGAN composed of 6 convolutional layers. The loss functions for the generator and discriminator are L1 and MSE, respectively. Random cropping of image patches with the size of 512 x 512 is implemented in the training process. For data standardization, a preprocessing step is performed to align all various mammograms into the physical pixel spacing of 0.1 mm. The epoch is empirically set to 100 for the training of CycleGAN, which can lead to the best mass detection performance in the validation dataset. For balanced training, each pass of style transfer, e.g., style A to style B, the training data of the source and target styles are set to the same number.

\begin{table*}
\caption{Performance comparison among MSVCL+ and other SOTA domain generalization methods w.r.t mass detection task. The assessment metric is mAP[\%].}\label{tab:comparison_detection}
\centering
{\fontsize{9}{11}\selectfont 
\begin{tabular}{c|c|c|c|c|c|c|c|c}
\hline
\multirow{2}{*}{Method}&\multicolumn{4}{c|}{Seen domain }&\multicolumn{4}{c}{Unseen domain} \\
\cline{2-9} & Style A & Style B & Style C & Avg. & Style D & Style E & Style F & Avg. \\ \hline
Baseline & 86.9 & 92.0 & 86.0 & 88.3$\pm$3.2 & 86.2 & 80.5 & 84.2 & 83.6$\pm$2.9 \\
BigAug  & 88.8 & 89.9 & 84.7 & 87.8$\pm$2.7 & 88.4 & 86.5 & 84.3 & 86.4$\pm$2.1  \\
DD  & 86.9 & 90.7 & 86.1 & 87.9$\pm$2.5 & 87.4 & 87.9 & 85.8 & 87.0$\pm$1.1\\
EISNet  & 89.3 & 89.1 & 82.8 & 87.1$\pm$3.7 & 86.2 & 83.8 & 85.6 & 85.2$\pm$1.2  \\
MSVCL &  91.8 & \textbf{94.0} & \textbf{89.1} & \textbf{91.6$\pm$2.5} & 87.3 & \textbf{89.7} & 88.4 & 88.5$\pm$1.2\\
MSVCL+(ours) & \textbf{92.9} & 93.8 & 87.0 & 91.2$\pm$3.7 & \textbf{90.4} & 89.4 & \textbf{94.1} & \textbf{91.3$\pm$2.5}\\
\hline
\end{tabular}
}
\end{table*}

\begin{table*}
\caption{Performance comparison among MSVCL+ and other SOTA domain generalization methods w.r.t. the four tasks. The assessment metrics for the four tasks are the same with table~\ref{tab:ablation}.}\label{tab:comparison}
\centering
{\fontsize{9}{11}\selectfont 
\begin{tabular}{c|c|c|c|c|c|c|c|c}
\hline
\multirow{2}{*}{Method}&\multicolumn{4}{c|}{Seen domain }&\multicolumn{4}{c}{Unseen domain} \\
\cline{2-9} & Detection & Match & BI-RADS & Density & Detection & Match & BI-RADS & Density\\ \hline
Baseline    & 88.3$\pm$3.2      & 80.0$\pm$6.7      & 87.4$\pm$6.7  & 91.7$\pm$2.5   & 83.6$\pm$2.9      & 85.0$\pm$7.4  & 83.3$\pm$18.4      & 76.3$\pm$16.3 \\
BigAug      & 87.8$\pm$2.7      & 88.9$\pm$3.2      & 89.0$\pm$4.4  & 90.0$\pm$4.0   & 86.4$\pm$2.1      & 87.9$\pm$6.6  & 83.3$\pm$17.0      & 78.7$\pm$5.0 \\
DD          & 87.9$\pm$2.5      & 84.4$\pm$4.8      & 87.5$\pm$8.3  & 90.3$\pm$5.5   & 87.0$\pm$1.1      & 88.5$\pm$2.7  & 85.3$\pm$14.2      & 82.0$\pm$6.2 \\
EISNet      & 87.1$\pm$3.7      & 86.8$\pm$2.7 & \textbf{89.6$\pm$4.7} & 88.7$\pm$7.1   & 85.2$\pm$1.2   & 90.9$\pm$2.8  & 82.4$\pm$17.0      & 80.7$\pm$7.6 \\
MSVCL & \textbf{91.6$\pm$2.5} & 89.6$\pm$4.3 & 89.3$\pm$4.9 & 91.7$\pm$3.2  & 88.5$\pm$1.2      & 91.5$\pm$2.9 & 85.9$\pm$9.4      & 81.0$\pm$4.4 \\
MSVCL+(ours) & 91.2$\pm$3.7    & \textbf{90.6$\pm$3.1}      & 89.0$\pm$5.5 & \textbf{92.0$\pm$2.6} & \textbf{91.3$\pm$2.5} & \textbf{92.9$\pm$2.5} & \textbf{87.8$\pm$9.3} & \textbf{84.3$\pm$8.3} \\
\hline
\end{tabular}
}
\end{table*}

The backbone model for the contrastive learning scheme and downstream tasks is ResNet-50. For fair comparison, the learning rate and batch size for all practices of the contrastive learning scheme are set at 0.3 and 256, respectively. Meanwhile, all contrastive learnings in all experiments share the same diversifying operations, including random cropping, random rotation in $\pm10^{\circ}$, horizontal flipping, and random color jittering (strength$=$0.2).

The model backbones of all downstream tasks are initialized with the pre-trained models from the self-learning scheme, i.e., stage (a) in Fig.~\ref{fig:overview}. For the training of FCOS models, the SGD optimization method is adopted with the parameters of learning rate, weight decay, and momentum set as $0.005$, $10^4$, and $0.9$, respectively. The epoch and batch size are set to 50 and 8, respectively, throughout all experiments. Several augmentation methods, e.g., random flipping and scaling, are also implemented in the training of FCOS.

For the training of the remaining downstream tasks, the SGD method is also employed, with the parameters of learning rate, weight decay, and momentum set as $0.001$, $10^5$, and $0.9$, respectively. The epoch and batch size are set to 50 and 128, respectively. For the tasks of mass matching and BI-RADS rating, the input of the model is the squared ROI derived from the box identified by the detection network. Specifically, the squared ROI is first defined by taking the largest side length of the detected box with the same center. To consider the context, the squared ROI is further enlarged by 20\%. For the task of breast density classification, the input to the model is the original mammograms. For these three tasks, the input images are resized to 224 x 224 pixels with the data augmentation of random flipping and scaling in the training process. The contrastive learning model for the mass matching task is trained with the setting of the hyper-parameter margin $m$ as 10 in Eq.~\ref{eq:img_blending}. To facilitate the downstream mass matching task, a nipple and muscle segmentation model is employed for the computation of object-to-nipple and object-to-chestwall distances.

\subsection{Ablation Study}

Table~\ref{tab:ablation_detection} reports the ablation study results on mass detection tasks w.r.t. different styles, and table~\ref{tab:ablation} shows the ablation study results of averaged performances of the four downstream tasks over the seen and unseen domains. The ablation experiments are considering three major aspects: 1) pre-trained models, 2) various combinative contrastive learning schemes on style and view domains, and 3) revised versions of contrastive learning compared to our previous work \citep{li2021domain}. Specifically, the performance comparison of different pre-trained models can be found in the first four rows of the Tables~\ref{tab:ablation_detection} and \ref{tab:ablation}. The first row stands for training from scratch, whereas the second row indicates performances with a pre-trained model from ImageNet. The third ``Mammo" row suggests that the SimCLR is trained from scratch with the unlabeled images of vendors A, B, and C. The pre-trained model of the fourth row, ``$\rm ImageNet\rightarrow Mammo$" is derived by the SimCLR, which was initialized with ImageNet parameters and tuned with the Mammo set used in the third rows. As can be seen, the pre-trained model from ``$\rm ImageNet\rightarrow Mammo$" derived by SimCLR is helpful for mass detection and other tasks.

\begin{figure*}[!t]
\centering
\makebox[\textwidth][c]{\includegraphics[width=1.0\textwidth]{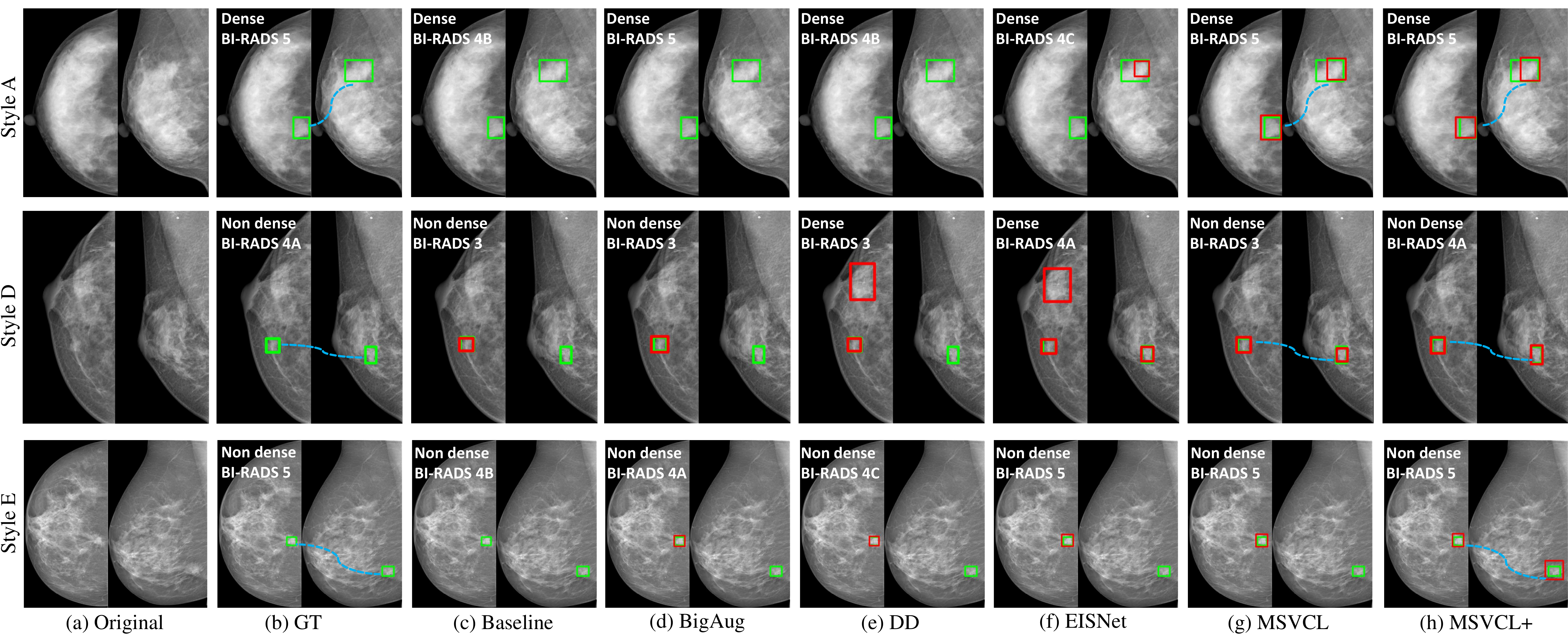}}
\caption{Visual comparison of downstream tasks results in different styles w.r.t. various implemented methods. The green boxes refer to the mass regions labeled by radiologists, while the red boxes indicate the mass regions detected by computerized methods. The blue dotted lines indicate the outcomes of mass matching. The BI-RADS rating and breast density classification results are annotated at the top of the CC views.}
\label{fig:detection_result} 
\end{figure*}

\begin{figure}[!t]
  \centering
  \includegraphics[width=0.48\textwidth]{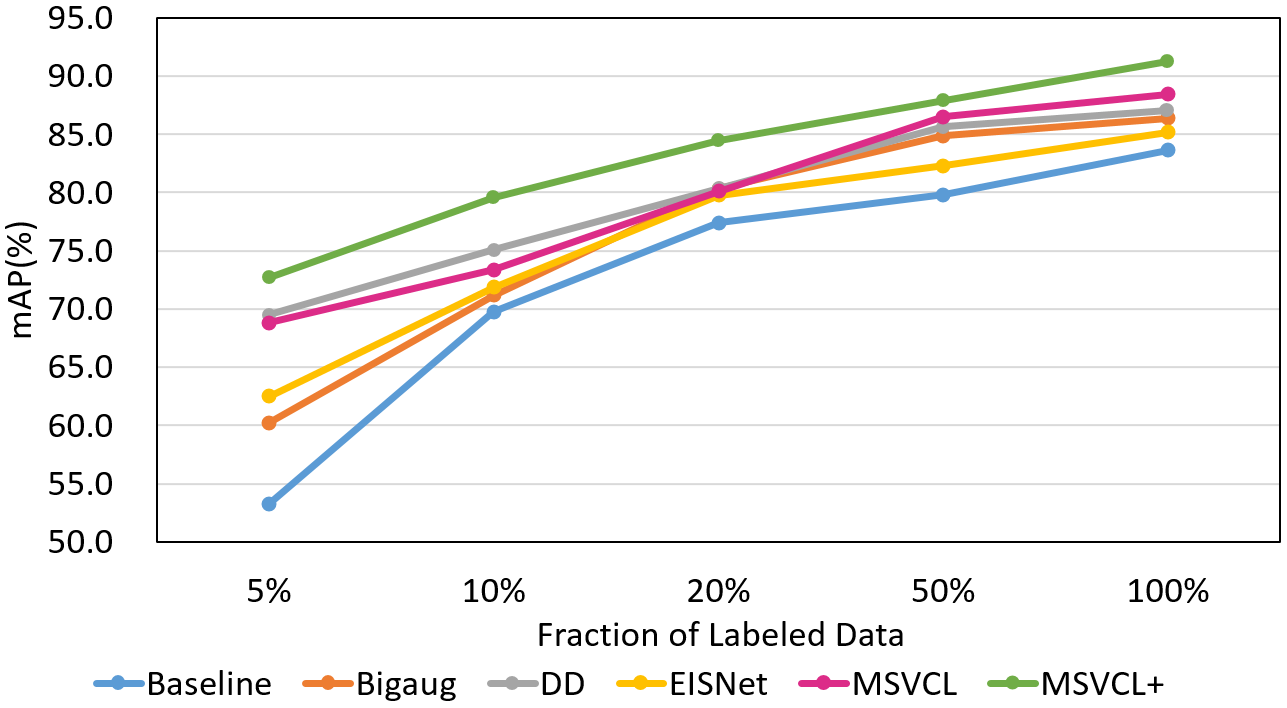}
  \caption{Performance comparison for all implemented domain generalization methods in data hungry experiments.}
  \label{fig:fraction}
\end{figure}

By observing the groups of $5^{th}$ to $7^{th}$ rows and $8^{th}$ to $10^{th}$ rows in Tables~\ref{tab:ablation_detection} and \ref{tab:ablation}, the performance can be mostly boosted by considering both style and view domains in the contrastive learning. Meanwhile, it may also be found that the pre-trained models from the revised versions, i.e., MSCL+, MVCL+, and MSVCL+, can mostly attain better performance on the five downstream tasks, particularly on unseen domains. The underlying reasons for the performance boost may be twofold. First, the increase in style diversity by the blending method may be helpful in seeking better embedding of style-invariant features. Second, the new selection strategy provides more reasonable and informative negative pairs for contrastive learning that may render the sought feature embedding a better representative for the mammographic image domain.

\subsection{Comparison with State-of-the-Art (SOTA) Methods}


To further compare with other domain generalization methods, three SOTA methods, i.e., BigAug \citep{zhang2020generalizing}, Domain Diversification (DD) \citep{kim2019diversify}, and EISNet \citep{wang2020learning} are implemented. The BigAug \citep{zhang2020generalizing} is a conventional data augmentation method, whereas the DD \citep{kim2019diversify} proposes a generative learning method for domain generalization. The EISNet \citep{wang2020learning} is a learning-based method that explores task-specific and domain-invariant features. Our method, on the other hand, can decouple the downstream tasks intrinsically and provide task- and domain-invariant features. For a fair comparison, the other methods, including BigAug, DD, and EISNet, were fine-tuned with the pre-trained backbone of SimCLR on ``$\rm ImageNet\rightarrow Mammo$", to ensure the use of the same amount of data. Table~\ref{tab:comparison_detection} reports the comparison results on the mass detection task with different styles, whereas Table~\ref{tab:comparison} presents the performance comparison for the four downstream tasks across both seen and unseen domains. The Baseline rows in Table~\ref{tab:comparison_detection} and~\ref{tab:comparison} suggest the results of SimCLR with ``$\rm ImageNet\rightarrow Mammo$". As can be found in Table~\ref{tab:comparison}, the MSVCL+ method achieves the best performance on unseen domains for the four downstream tasks.

A data-hungry experiment is also conducted for the comparison of SOTA methods. Five fraction settings, i.e., 5\%, 10\%, 20\%, 50\%, and 100\% of labeled (training) data, are applied for the experiment to illustrate the learning capability of domain generalization methods in terms of the amount of training data. The data-hungry experiment is conducted on the detection task, and the corresponding results are shown in Fig.~\ref{fig:fraction}. As can be observed, the MSVCL+ outperforms other methods even with a very small amount of data, i.e., the 5\% fraction setting. Fig.~\ref{fig:detection_result} lists several results of the implemented downstream tasks from different domain generalization methods for visual comparison. It can be observed that MSVCL+ can help the tasks attain better performance. 


\section{Conclusion}
A novel domain generalization method, denoted as MSVCL+, is proposed to endow the deep learning models with better generalizability to the vendor style and view domains. The MSVCL+ has been shown to be helpful for four mammographic image analysis tasks, i.e., detection, matching, BI-RADS rating, and breast density classification. The MSVCL+ can provide more robust feature embedding against various vendor-style domains. Experimental results suggest that our method can effectively improve mammographic image analysis tasks on unseen domains compared to the four implemented SOTA domain generalization methods. In particular, our method achieves the best performance on the public datasets INbreast and DDSM, where the domain gaps may not only include image styles but also population factors. The INbreast and DDSM datasets were collected from Western women. In contrast, our methods were trained on mammograms acquired from Asian women. Therefore, the generalization efficacy of our proposed method is further corroborated. 

\section*{Acknowledgments}
This work was supported in part by the Key Research and Development Program of Guangdong Province, China, under Grant 2021B0101420006; in part by Guangdong Provincial Key Laboratory of Artificial Intelligence in Medical Image Analysis and Application under Grant 2022B1212010011; in part by the National Natural Science Foundation of China under Grant 82171920; and in part by the National Natural Science Foundation of China under Grant 82071878.

\bibliographystyle{model2-names.bst}\biboptions{authoryear}
\bibliography{refs}



\end{document}